\documentclass{preprint}
\usepackage{graphicx}
\usepackage[colorlinks=true]{hyperref}
\usepackage[super,comma,sort&compress]{natbib}
\usepackage{xcolor}
\usepackage{booktabs}
\usepackage{xurl}
\usepackage[inline]{enumitem}

\title{LLM assisted writing deserves empirical evaluation}
\author[1]{Xuan Zhong Feng}
\author[1]{Yi Lin}
\author[1]{Yiye Zhang}
\author[2]{Chunhua Weng}
\author[1,*]{Yifan Peng}
\affil[1]{Department of Population Health Sciences, Weill Cornell Medicine, New York, NY, USA}
\affil[2]{Department of Biomedical Informatics, Columbia University, New York, NY, USA}
\affil[*]{Corresponding author(s). Email(s): \url{yip4002@med.cornell.edu}}

\begin{document}

\maketitle

\begin{abstract}
LLM-assisted writing is often treated as a detection problem, as it raises questions about clarity, integrity, equity, and evaluation. An analysis of 69,209 Health Informatics papers links it to more focused presentation, broader citation practices, and more globally distributed authorship. These patterns do not prove better science, but they support evaluating manuscripts by scholarly quality and accountability rather than by tool use.
\end{abstract}

\section*{Main text}

\subsection*{Introduction}

Large language models (LLMs) have quickly become part of ordinary scientific work. Researchers use them to search, summarize, draft, edit, translate, and reorganize text\cite{fecher2025friend, asai2026synthesizing}. Academic debate has often focused on whether LLM-assisted writing can be detected, disclosed, restricted, or separated from ``real'' scholarship. While concerns about transparency, authorship, fabricated content, bias, and overreliance are important\cite{kobak2025delving}, LLM-assisted writing may also alter the practical conditions under which researchers communicate, especially for authors working across languages, institutions, and disciplines. Their role in scientific writing is therefore an empirical and social question, not only a technical one. Treating LLM-assisted writing mainly as a detection problem risks missing how it may reshape clarity, citation practices, authorship patterns, and recognition.

The question is not simply whether LLM-assisted writing is beneficial or harmful.
Scientific writing has long depended on many forms of assistance, such as mentors, collaborators, professional editors, translators, reporting guidelines, and journal templates. LLM-assisted writing adds a new form of assistance to this ecosystem. It differs from earlier forms because it can generate fluent text at scale, summarize literature, translate across languages, and introduce errors or unsupported claims in ways that may be difficult to detect. Their effects are likely to depend on the task, the user, the field, the institutional setting, and the level of human oversight.

\begin{figure}[p]
\centering
\includegraphics[width=\textwidth]{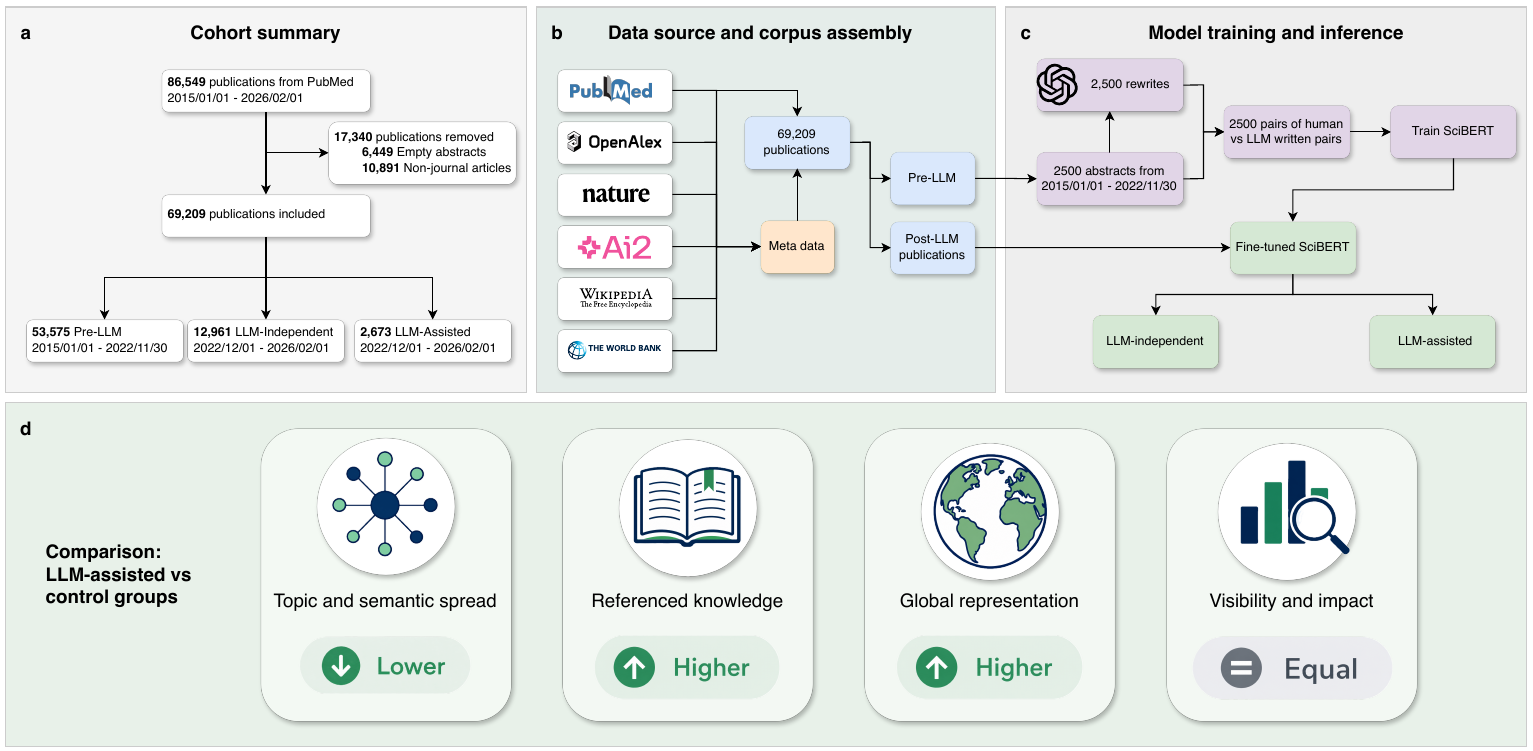}
\caption{\textbf{Summary of observed patterns.}
\textbf{a--c,} PubMed-indexed Health Informatics publications were assembled with metadata from external sources and grouped into pre-LLM, LLM-independent, and LLM-assisted papers using a SciBERT classifier trained on human-written and LLM-rewritten abstracts.
\textbf{d,} Compared with control groups, LLM-assisted papers showed lower topic and semantic spread, higher referenced-knowledge diversity, higher global representation, and similar visibility and impact.}
\label{fig:study_design}
\end{figure}

To examine how LLM-assisted writing is associated with scientific communication, we analyzed 69,209 PubMed-indexed Health Informatics articles. We grouped papers as pre-LLM or post-LLM, using the public release of ChatGPT in late 2022 as a practical boundary\cite{openai2022chatgpt}, and classified post-LLM papers into LLM-independent and LLM-assisted using a SciBERT classifier\cite{beltagy-etal-2019-scibert}. We then fine-tuned the SciBERT to distinguish original abstracts from LLM-rewritten versions and selected a conservative classification threshold to prioritize precision, as detailed in the methods documentation linked in Code availability. We then compared bibliometric, semantic, reference, and authorship metadata across groups (Figure~\ref{fig:study_design}). The results suggest that LLM-assisted writing is not simply a marker of lower-quality or less serious scholarship. Instead, it was associated with a more focused presentation, more comprehensive citations of related work, and broader researcher participation.

Below, we consider these patterns in turn, covering semantic concentration, reference breadth, authorship and global participation, and visibility and recognition, before distinguishing LLM-assisted writing from agentic research.

\subsection*{Semantic concentration in scholarly writing}

\begin{figure}[p]
\centering
\includegraphics[width=\textwidth,height=0.68\textheight,keepaspectratio]{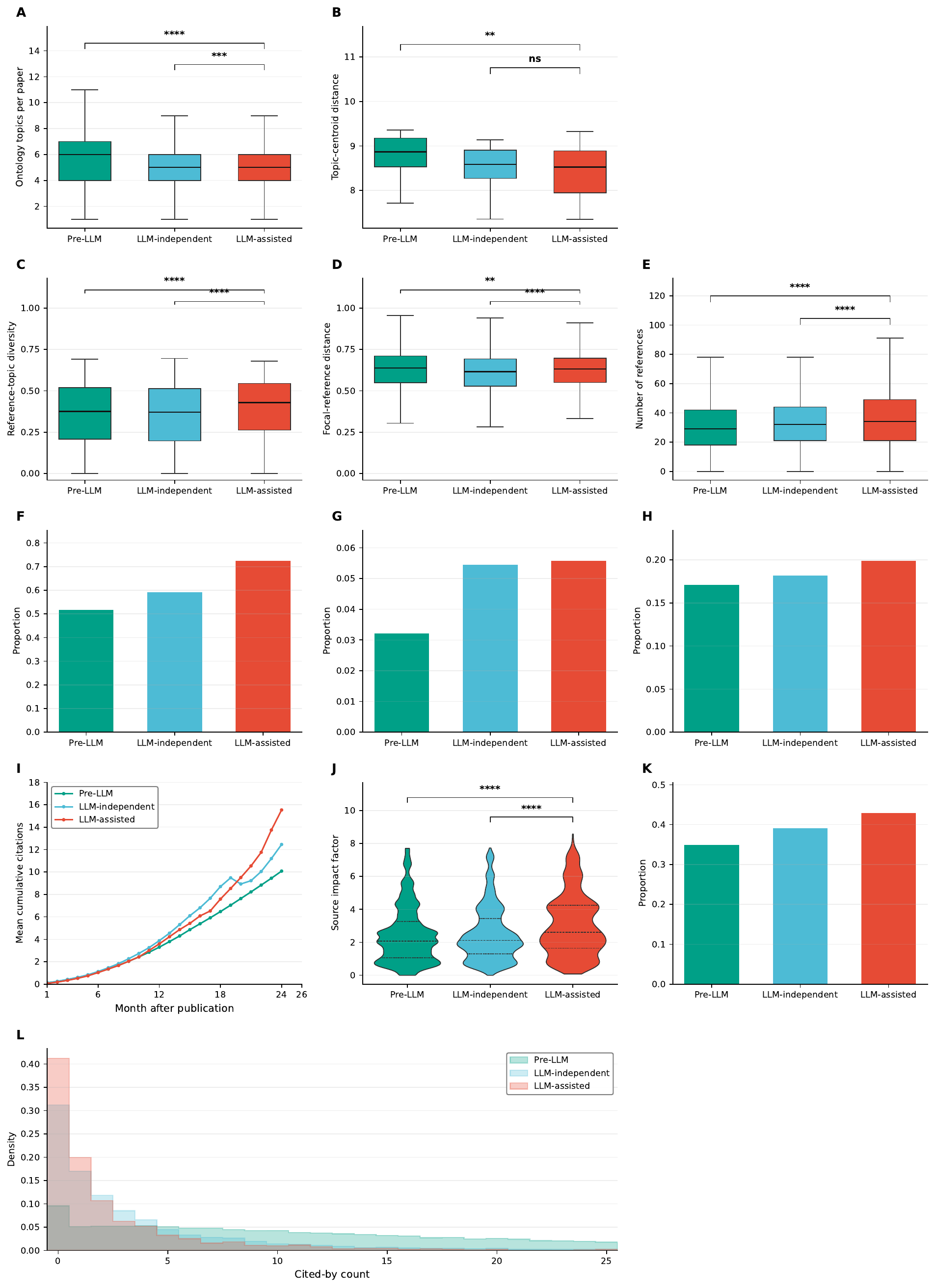}
\caption{\textbf{Focal topical breadth and semantic concentration, reference-list size and topical diversity of cited knowledge, linguistic, economic, and international collaboration characteristics of authorship, and publication visibility and citation impact across analytic groups.}
\textbf{a,} Ontology topics assigned per paper.
\textbf{b,} Mean Euclidean distance from each paper to the centroid of its assigned focal topic or topics. Lower values indicate greater semantic concentration around topic centers.
\textbf{c,} Rao-Stirling diversity of reference topics.
\textbf{d,} Rao-Stirling distance of focal-reference.
\textbf{e,} Number of references per paper.
\textbf{f,} Proportion of papers with first-author affiliation in a non-Anglophone setting.
\textbf{g,} Proportion of papers with last-author affiliation in a low- or middle-income economy.
\textbf{h,} Proportion of papers with intercontinental collaboration.
\textbf{i,} Cumulative citations within 24 months.
\textbf{j,} Journal impact factor.
\textbf{k,} Proportion of papers published in Gold Open Access journals.
\textbf{l,} Cited-by count distribution.
Statistical annotations indicate pairwise Mann--Whitney U tests.
ns: $p>0.05$, *: $p < 0.05$, **: $p < 0.01$, ***: $p < 0.001$, ****: $p < 0.0001$.}
\label{fig:combined_results}
\end{figure}

LLM-assisted papers appeared more focused in their topical and semantic presentation than both post-era LLM-independent and pre-LLM papers (Figure~\ref{fig:combined_results}a-b). They were assigned fewer topics per paper, and their title-and-abstract embeddings were closer to the semantic centroids of those topics. Although a similar trend was observed across broader post-LLM papers, the pattern was most pronounced among the LLM-assisted group.

This concentration admits multiple interpretations. On the one hand, it may reflect a clearer positioning of a manuscript within an established research area. A more focused presentation can help readers understand a paper's contribution, situate it within existing work, and assess its relevance. By encouraging language that aligns with disciplinary conventions, LLM-assisted writing may improve clarity, discoverability, and accessibility, particularly for authors writing across linguistic or institutional boundaries.

On the other hand, LLM-assisted writing may reduce rhetorical diversity and make distinctions between papers less apparent. Importantly, semantic closeness to the language of a field does not by itself indicate either novelty or lack of novelty. A paper can be semantically close to field norms and still contain a novel method, dataset, application, or empirical result; it can also be fluently written while making a limited contribution.
Accordingly, the question is not whether LLM-assisted scientific writing becomes more concentrated, but whether the resulting manuscripts continue to communicate specific, accurate, and substantively original contributions.

\subsection*{Reference patterns and knowledge breadth}

The reference patterns point to another area of potential change. LLM-assisted papers cited more references than post-era LLM-independent papers and showed higher reference-topic diversity (Figure~\ref{fig:combined_results}c--e). Their bibliographies also drew from a broader range of topics.

This contrast matters because LLMs are used for more than polishing prose. Researchers increasingly use them as writing assistants to summarize literature, identify adjacent concepts, generate search terms, and reorganize related work\cite{fecher2025friend,asai2026synthesizing}. If these practices lower the cost of exploring neighboring literatures, they may expand the range of papers that authors consider relevant. 

However, broader retrieval is not synonymous with better selection. Without clear standards for assessing \emph{relevance}, access to a larger set of candidate references may increase both meaningful interdisciplinary connections and the inclusion of citations that are only weakly related to the scientific argument.
A more extensive and diverse bibliography, therefore, does not necessarily indicate deeper synthesis. Citations can be superficial, misplaced, or only loosely connected to the claims they support. 

Such limitations are not unique to LLM-assisted writing. Human-written papers can likewise contain decorative citations and incomplete literature reviews. Depending on how they are used and verified, LLMs may amplify, reduce, or leave such existing practices unchanged.
Developing automated methods to evaluate citation accuracy, balance, and relevance may become increasingly important as LLM-assisted writing becomes more common.

\subsection*{Authorship and global participation}

The biggest observed difference was in the authorship context. Compared with both post-era LLM-independent papers and pre-LLM papers, LLM-assisted papers had a higher share of first authors affiliated with non-Anglophone settings. They also showed modestly higher representation from low- and middle-income economies and a higher level of intercontinental collaboration (Figure~\ref{fig:combined_results}f--h).

One plausible interpretation is that LLMs may help reduce some barriers to participation in English-language scientific publishing. English remains the dominant language of indexed biomedical research, and researchers working outside Anglophone settings often face additional costs for editing, translation, and adherence to international publication norms. LLMs may reduce these constraints by providing on-demand language assistance and by helping authors align manuscripts with expected scientific genres.

However, the potential benefits of LLM-assisted writing may not be distributed evenly. Access to advanced writing assistants varies across researchers and institutions, and unequal access may introduce new forms of advantage. If these tools substantially reduce the time required for literature review, drafting, revision, and manuscript preparation, researchers with greater access to such resources may be able to produce and submit work more rapidly. In competitive publication environments, these productivity advantages could accumulate over time, potentially widening disparities between researchers and institutions with differing levels of access. Conversely, scholars who rely primarily on traditional writing processes may face increasing pressure to compete on both speed and scientific quality. 
Consequently, although LLMs may lower some existing barriers to publication, their overall effect on equity in scientific participation remains uncertain and will likely depend on how access to these tools is distributed across the research community.

\subsection*{Visibility and recognition}

A common concern is that LLM-assisted writing may produce lower-quality work that receives less recognition. Our analyses showed no consistent pattern of reduced visibility or recognition for LLM-assisted papers. LLM-assisted papers had lower raw cited-by counts, but these counts are sensitive to publication timing and to the time available for citations to accumulate. Comparisons within fixed post-publication windows showed no consistent citation advantage or disadvantage compared to LLM-independent papers, despite their longer reference lists.

Journal publishing patterns also differed between groups. LLM-assisted papers were published in journals with higher average impact factor and were more likely to appear in Gold Open Access journals, although subscription/hybrid journals remained the predominant publication venue in both groups.

These findings speak to visibility and recognition, not intrinsic scientific quality. Citation counts, journal impact factors, and journal access modes are imperfect indicators of scientific value, as they are shaped by publication timing, field norms, editorial selection, and topic popularity.

Within these limitations, the observed patterns do not suggest reduced early recognition for LLM-assisted work. Whether these patterns are robust across disciplines, time horizons, and modes of LLM use as writing assistance remains an open question.

\subsection*{LLM-assisted writing vs agentic research}

This study provides a large-scale empirical assessment of LLM-assisted scientific writing. A key conceptual distinction is between LLM-assisted writing and human intellectual collaboration, such as co-authorship or co-research. Here, we focus on LLM-assisted writing that does not generate scientific intent, formulate research questions, design studies, or assume responsibility for the validity of results. They function as non-agentic tools for language transformation, summarization, and restructuring rather than as contributors to epistemic decision-making. In this sense, LLM assistance is more closely analogous to writing support systems, statistical software, or editorial services than to intellectual partnership. However, in practice, the extensive use of LLMs for ideation, literature synthesis, or analytical support may blur the boundaries between writing assistance and research assistance. This ambiguity reinforces the importance of transparent disclosure and careful attribution, while maintaining a conceptual separation between tools that support expression and agents that bear scientific responsibility.

Overall, LLM-assisted writing should be understood neither as inherently detrimental nor inherently beneficial to scientific communication. 
Rather, it represents a shifting component of the scholarly production process whose consequences depend on how it is adopted, regulated, and integrated into existing academic practices.
Our analysis cannot determine whether semantic concentration increases over time, leaving this question for future longitudinal research.
More broadly, debates about LLM-assisted writing should move from detection and moral classification toward evaluating how it changes communication quality, citation practices, access, and accountability.
Future work should prioritize more robust and fine-grained measurement of LLM use in scientific writing, including variation in intensity, purpose (e.g., editing versus drafting), and stage of manuscript development. Future research should also examine whether LLM use increases submission volume and places additional demands on editorial assessment and peer review, and whether editors and reviewers adopt LLM tools in response to these demands.
In the meantime, editorial and research evaluation should focus less on whether LLM assistance was used and more on whether the resulting scholarship is accurate, transparent, well-supported, and intellectually accountable.

\section*{Data availability}

All data used in this study were obtained from PubMed and OpenAlex and are publicly available.

\section*{Code availability}

The complete methods documentation is available at \url{https://github.com/bionlplab/llm-assisted-writing-methods}. The repository provides a readable methods document, a PDF, and the corresponding LaTeX source.

\section*{Declaration statements}

\subsection*{1. Acknowledgements}

This research was supported by the National Library of Medicine under grant numbers R01LM014306, R01LM014344, and R01LM014573 awarded to Yifan Peng. The funder had no role in study design, data collection and analysis, manuscript preparation, or the decision to publish.

\subsection*{2. Author Contributions}
Study concepts/study design, X.Z.F., Y.L., Y.P.; manuscript drafting or manuscript revision for important intellectual content, X.Z.F., Y.P.; agrees to ensure any questions related to the work are appropriately resolved, X.Z.F., Y.L., Y.P.; literature research, X.Z.F., Y.L., Y.P.; experimental studies, X.Z.F., Y.L., Y.P.; data interpretation and statistical analysis, X.Z.F., Y.L., Y.P.; and manuscript editing, X.Z.F., Y.L., Y.Z., C.W., Y.P.; read and approval of final version of the submitted manuscript, X.Z.F., Y.L., Y.Z., C.W., and Y.P.

\subsection*{3. Competing Interests}\label{competing_interests}
The authors declare no competing financial or non-financial interests.

\clearpage
\bibliographystyle{naturemag}
\bibliography{ref}

\clearpage

\clearpage

\end{document}